\documentclass{article}

\usepackage{microtype}
\usepackage{graphicx}
\usepackage{subfigure}
\usepackage{booktabs} 

\usepackage{hyperref}

\usepackage[accepted]{icml2024}

\usepackage{amsmath}
\usepackage{amssymb}
\usepackage{mathtools}
\usepackage{amsthm}

\usepackage[capitalize,noabbrev]{cleveref}

\theoremstyle{plain}
\newtheorem{theorem}{Theorem}[section]
\newtheorem{proposition}[theorem]{Proposition}

\theoremstyle{definition}

\theoremstyle{remark}

\newcommand{\ie}{\textit{i}.\textit{e}.}

\newcommand{\bI}{\mathbf{I}}

\newcommand{\hbY}{\hat{\mathbf{Y}}}
\newcommand{\bY}{\mathbf{Y}}
\newcommand{\R}{\mathbb{R}}
\newcommand{\bC}{\mathbf{C}}
\newcommand{\bL}{\mathbf{L}}
\newcommand{\bF}{\mathbf{F}}
\newcommand{\bP}{\mathbf{P}}
\newcommand{\bZ}{\mathbf{Z}}
\newcommand{\N}{\mathcal{N}}
\newcommand{\bW}{\mathbf{W}}
\newcommand{\bx}{\mathbf{x}}
\newcommand{\bX}{\mathbf{X}}

\def\figref#1{Figure~\ref{#1}}
\def\tabref#1{Table~\ref{#1}}
\def\eqref#1{Eq.~\ref{#1}}

\DeclareMathOperator{\UP}{UP}
\DeclareMathOperator{\GAP}{GAP}
\DeclareMathOperator{\E}{E}
\DeclareMathOperator{\Var}{Var}
\DeclareMathOperator{\ReLU}{ReLU}
\DeclareMathOperator{\BatchNorm}{BatchNorm}
\DeclareMathOperator{\fuse}{fuse}
\DeclareMathOperator{\ScaleEqualizer}{ScaleEqualizer}

\usepackage{physics}  

\usepackage{listings}
\usepackage{xcolor}
\definecolor{codegreen}{rgb}{0,0.6,0}
\definecolor{codegray}{rgb}{0.5,0.5,0.5}
\definecolor{codepurple}{rgb}{0.58,0,0.82}
\definecolor{backcolour}{rgb}{0.95,0.95,0.92}

\lstdefinestyle{mystyle}{
    backgroundcolor=\color{backcolour},   
    commentstyle=\color{codegreen},
    keywordstyle=\color{magenta},
    numberstyle=\tiny\color{codegray},
    stringstyle=\color{codepurple},
    basicstyle=\ttfamily\footnotesize,
    breakatwhitespace=false,         
    breaklines=true,                 
    captionpos=b,                    
    keepspaces=true,                 
    numbers=left,                    
    numbersep=5pt,                  
    showspaces=false,                
    showstringspaces=false,
    showtabs=false,                  
    tabsize=2
}

\makeatletter
\renewcommand{\paragraph}{%
  \@startsection{paragraph}{4}%
  {\z@}{0.00ex \@plus 1ex \@minus .2ex}{-1em}%
  {\normalfont\normalsize\bfseries}%
}
\makeatother

\usepackage[textsize=tiny]{todonotes}

\icmltitlerunning{Scale Equalization for Multi-Level Feature Fusion}

\begin{document}

\twocolumn[
	\icmltitle{Scale Equalization for Multi-Level Feature Fusion}




	\begin{icmlauthorlist}
		\icmlauthor{Bum Jun Kim}{yyy}
		\icmlauthor{Sang Woo Kim}{yyy}
	\end{icmlauthorlist}

	\icmlaffiliation{yyy}{Department of Electrical Engineering, Pohang University of Science and Technology, Pohang, South Korea}

	\icmlcorrespondingauthor{Sang Woo Kim}{swkim@postech.edu}

	\icmlkeywords{Machine Learning, ICML}

	\vskip 0.3in
]



\printAffiliationsAndNotice{}  

\begin{abstract}
	Deep neural networks have exhibited remarkable performance in a variety of computer vision fields, especially in semantic segmentation tasks. Their success is often attributed to multi-level feature fusion, which enables them to understand both global and local information from an image. However, we found that multi-level features from parallel branches are on different scales. The scale disequilibrium is a universal and unwanted flaw that leads to detrimental gradient descent, thereby degrading performance in semantic segmentation. We discover that scale disequilibrium is caused by bilinear upsampling, which is supported by both theoretical and empirical evidence. Based on this observation, we propose injecting scale equalizers to achieve scale equilibrium across multi-level features after bilinear upsampling. Our proposed scale equalizers are easy to implement, applicable to any architecture, hyperparameter-free, implementable without requiring extra computational cost, and guarantee scale equilibrium for any dataset. Experiments showed that adopting scale equalizers consistently improved the mIoU index across various target datasets, including ADE20K, PASCAL VOC 2012, and Cityscapes, as well as various decoder choices, including UPerHead, PSPHead, ASPPHead, SepASPPHead, and FCNHead.
\end{abstract}

\section{Introduction}
\label{sec:intro}
Deep neural networks have shown remarkable performance, especially in the computer vision field. Their substantial modeling capability has enabled us to develop significantly accurate models with rich image features for a wide range of vision tasks, including object detection and semantic segmentation.

One challenge in computer vision tasks is understanding both the global and local contexts of an image \citep{DBLP:conf/iclr/ReddiKK18,DBLP:conf/cvpr/TuTZYMBL22}. Indeed, the cascade architecture of a deep neural network faces difficulty in understanding multiple contexts of an image owing to the single-level feature it uses. To address this problem, modern vision networks have employed a parallel architecture that aggregates multi-level features in different spatial sizes to extract both global and local information from an image. For semantic segmentation as an example, multi-level feature fusion has been adopted in numerous models such as UPerNet \citep{DBLP:conf/eccv/XiaoLZJS18}, PSPNet \citep{DBLP:conf/cvpr/ZhaoSQWJ17}, DeepLabV3 \citep{DBLP:journals/corr/ChenPSA17}, DeepLabV3+ \citep{DBLP:conf/eccv/ChenZPSA18}, FCN \citep{DBLP:conf/cvpr/LongSD15}, and U-Net \citep{DBLP:conf/miccai/RonnebergerFB15}.

Although the parallel architecture builds multiple fastlanes to facilitate multi-level features to contribute to output, if certain features are not involved in the fusion, they simply waste computational resources. Initially, all feature branches should be exploited to explore their potential usefulness, and after training, their optimal combination should be obtained. Thus, the underlying assumption of multi-level feature fusion is that, at least in an initialized state, all multi-level features will participate in producing a fused feature. However, we claim that existing architectural design for multi-level feature fusion has a potential problem of scale disequilibrium, which yields unwanted bias that diminishes the contribution of several features. Specifically, the multi-level features exhibit different scales at initialization, which leads to different contributions and gradient scales, thereby hindering training with gradient descent. We identify the cause of the scale disequilibrium---bilinear upsampling, which is used to enlarge multi-level features to the same spatial size. Demonstration of the scale disequilibrium caused by bilinear upsampling is provided theoretically and empirically.

To solve the scale disequilibrium problem, this study proposes injecting scale equalizers into multi-level feature fusion. The scale equalizer normalizes each feature using the global mean and standard deviation, which guarantees scale equilibrium across multi-level features. Because the proposed scale equalizer is simply global normalization, which uses empirical values for subtraction and division, its implementation is easy and hyperparameter-free, requires little extra computation that is actually free, and assures scale equilibrium for any datasets and architectures. Experiments on semantic segmentation tasks showed that applying scale equalizers for multi-level feature fusion consistently improved the mIoU index across extensive experimental setups, including datasets and backbones.

\section{Background}
\label{sec:back}
\paragraph{Formulation} This study considers the standard framework for supervised learning of semantic segmentation networks because it is a representative task using multi-level feature fusion. Let $\bI \in \R^{H \times W \times C}$ be an input image to a semantic segmentation network, where $H, W$ is the size of the image and $C$ represents the number of image channels. The objective of semantic segmentation is to generate a semantic mask $\hbY \in \R^{H \times W \times N_c}$ that classifies each pixel in the image $\bI$ into one of the $N_c$ categories. A deep neural network, which comprises an encoder and decoder, is employed as a semantic segmentation model that outputs $\hbY$ from the input image $\bI$. The encoder is a backbone network with several stages where the input image first goes through. The decoder---also referred to as the head---uses a set of intermediate feature maps $\{\bC_i\}$ from the encoder to produce the segmentation output $\hbY$. To quantify a difference between the prediction $\hbY$ and the ground truth $\bY$, a loss function such as pixel-wise cross-entropy is used. With gradient descent optimization for the loss function, the encoder and decoder are trained together on an image-label pair dataset by the fine-tuning strategy, where the encoder begins with a pretrained weight whereas the decoder is trained from scratch.

\subsection{Multi-Stage Feature Fusion}
The last feature map of the encoder contains rich, high-level information on the image \citep{DBLP:journals/pami/ChenPKMY18} and is included in the set of feature maps used by the decoder. However, each stage of the encoder yields a downsampled feature map. Thus, the last feature map exhibits a severe downsampling rate, losing fine details in the image \citep{DBLP:journals/corr/ChenPSA17}. To address this problem, modern decoders have used multiple feature maps from several stages to aggregate information with various spatial sizes \citep{DBLP:conf/cvpr/KirillovGHD19,DBLP:conf/cvpr/ZhengLZZLWFFXT021}. We refer to this practice as \emph{multi-stage feature fusion}. For multi-stage feature fusion, the common choice on the set of features is to use the encoder outputs of the second to fifth stages, \ie, $\{\bC_2, \bC_3, \bC_4, \bC_5\}$ \citep{DBLP:conf/cvpr/LinDGHHB17}. The use of the first stage output is commonly avoided because it requires large GPU memory. For convolution-based residual networks \citep{DBLP:conf/cvpr/HeZRS16} and certain vision transformers such as Swin \citep{DBLP:conf/iccv/LiuL00W0LG21}, the downsampling ratios of the four encoder features are $\{4, 8, 16, 32\}$. Others, such as vanilla vision transformers \citep{DBLP:conf/iclr/DosovitskiyB0WZ21}, keep the same spatial size for the four encoder features. Despite the promising results of the latter, in general vision tasks, progressive downsampling is critical to encouraging heterogeneous characteristics in multiple feature maps, which advocates the former networks. This study targets the former and describes a problem regarding feature fusion using different downsampling ratios. The remainder of this section reviews the detailed mechanism of a modern decoder with multi-stage feature fusion.

\paragraph{UPerHead} UPerHead, the head deployed in UPerNet \citep{DBLP:conf/eccv/XiaoLZJS18}, is a prime example of a decoder designed for multi-stage feature fusion. Recent vision transformers have preferred the UPerHead for semantic segmentation tasks \citep{DBLP:conf/iclr/DosovitskiyB0WZ21,DBLP:conf/iclr/Bao0PW22,DBLP:conf/cvpr/HeCXLDG22}, and their remarkable performance let it be one of the most widely used decoders in the current state. The UPerHead comprises different modules, such as the pyramid pooling module (PPM) \citep{DBLP:conf/cvpr/ZhaoSQWJ17}, feature pyramid network (FPN) \citep{DBLP:conf/cvpr/LinDGHHB17}, and convolutional unit block, which is composed of convolution, batch normalization, and ReLU operation. Firstly, each of the three feature maps $\{\bC_2, \bC_3, \bC_4\}$ is subjected to a convolutional unit block to yield laterals $\{\bL_2, \bL_3, \bL_4\}$. Additionally, the last lateral $\bL_5$ is produced from the last feature map $\bC_5$ using the PPM module that is described in \cref{sec:singlestage}. Now, a subnetwork called FPN performs the top-down pathway to laterals to obtain its output $\bF_i = \bL_i + \UP_2(\bF_{i+1})$ for $i \in \{2, 3, 4\}$ and $\bF_5 = \bL_5$, where the operation $\UP_r$ denotes $r \times$ bilinear upsampling. Subsequently, FPN applies a convolutional unit block $h_i$ to each result to yield its output $\bP_i = h_i(\bF_i)$ for $i \in \{2, 3, 4, 5\}$. The set of FPN output $\{\bP_2, \bP_3, \bP_4, \bP_5\}$ has the same spatial size as encoder features $\{\bC_2, \bC_3, \bC_4, \bC_5\}$, keeping their downsampling ratios $\{4, 8, 16, 32\}$. Thus, feature fusion for FPN outputs requires $2^{i-2} \times$ bilinear upsampling for each $\bP_i$ to ensure that they share the same spatial size of $H/4 \times W/4$. Finally, they can be concatenated together with respect to channel dimension and fused with a convolutional unit block $h$ as
\begin{align}
	\bZ = h([\bP_2; \UP_2(\bP_3); \UP_4(\bP_4); \UP_8(\bP_5)]).
\end{align}
The fused feature $\bZ$ is then subjected to a $1\times1$ convolution and a $4\times$ bilinear upsampling to yield a predicted semantic mask $\hbY$, which has the size of $H \times W \times N_c$.

\begin{figure*}[t]
	\begin{center}
		\centerline{\includegraphics[width=1.0\textwidth]{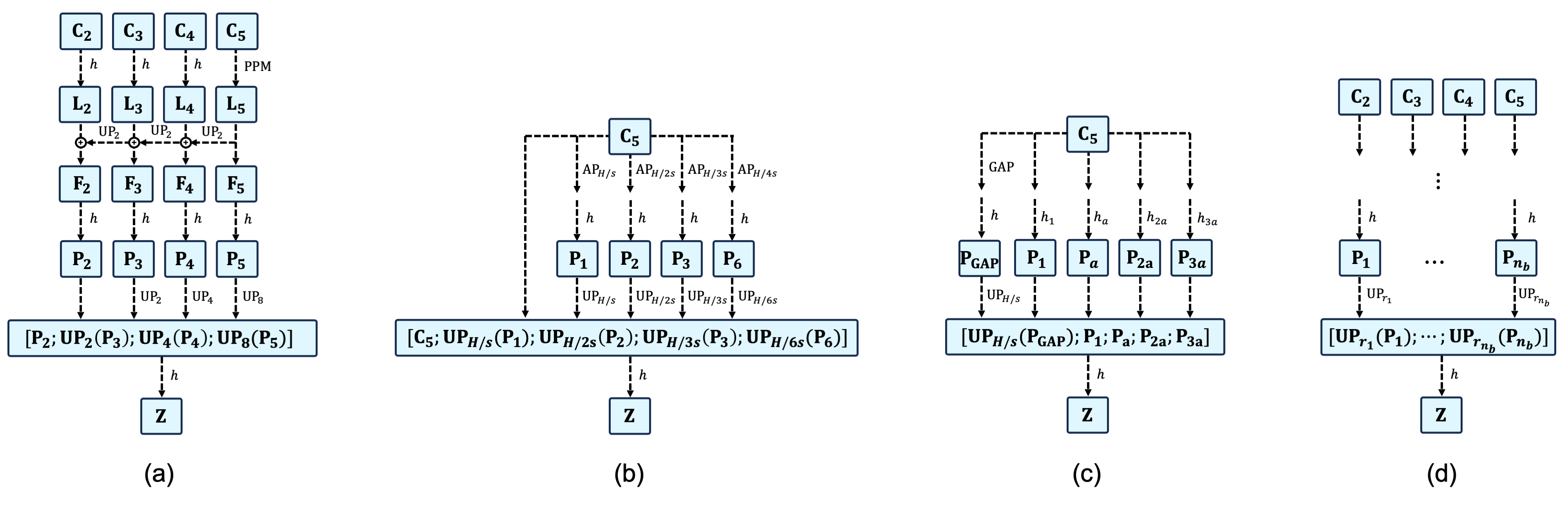}}
		\caption{
			Visualization of the architecture of modern decoders: (a) UPerHead, (b) PSPHead, (c) ASPPHead and SepASPPHead, and (d) their general form.
		}
		\label{fig:head}
	\end{center}
	\vskip -0.2in
\end{figure*}

\subsection{Single-Stage Feature Fusion}
\label{sec:singlestage}
Although multi-stage feature fusion uses a set of encoder features from several stages, certain decoders such as PSPHead \citep{DBLP:conf/cvpr/ZhaoSQWJ17} or ASPPHead \citep{DBLP:journals/corr/ChenPSA17} only use a single feature map from the last stage $\bC_5$. They modify the encoder to exhibit a downsampling ratio of 8 or 16 at the last stage, which is referred to as the output stride. Denoting the output stride as $s$, the spatial size of the last feature map $\bC_5$ is $(H/s) \times (W/s)$. These decoder heads perform dissimilar feature fusion: From a single-stage feature, multiple feature maps with various sizes are produced, which are then fused. We refer to this type of feature fusion as \emph{single-stage feature fusion}. In a similar but different way, single-stage feature fusion enables the decoder to extract both global and local contexts from the targeted feature map. The remainder of this section reviews the detailed mechanisms of modern decoders with single-stage feature fusion.

\paragraph{PSPHead} PSPHead refers to the head deployed in PSPNet \citep{DBLP:conf/cvpr/ZhaoSQWJ17}. Its underlying mechanism is to extract global and local contexts from a feature map using multiple branches, which is referred to as a pyramid pooling module (PPM). Targeting the last feature map $\bC_5$, it performs four average poolings in parallel, which yields features with spatial sizes $1 \times 1$, $2 \times 2$, $3 \times 3$, and $6 \times 6$. Subsequently, a convolutional unit block is applied to each branch, and then each result is upsampled to fit the size of $\bC_5$. Along with the feature map $\bC_5$, the results from the four branches are concatenated together with respect to channel dimension. Finally, a convolutional unit block $h$ is applied to fuse them. Denoting the outputs of convolutional unit blocks in parallel branches as $\{\bP_1, \bP_2, \bP_3, \bP_6\}$, fusing them is represented as
\begin{align}
	\bZ = h([\bC_5; & \UP_{H/s}(\bP_1); \UP_{H/2s}(\bP_2);    \\
	                & \UP_{H/3s}(\bP_3); \UP_{H/6s}(\bP_6)]),
\end{align}
where $H = W$ is assumed for notational simplicity. Similarly, the fused feature $\bZ$ is then subjected to a $1\times1$ convolution and a $4\times$ bilinear upsampling to yield a predicted semantic mask $\hbY$, which has the size of $H \times W \times N_c$.

\paragraph{ASPPHead and Others} ASPPHead refers to the head deployed in DeepLabV3 \citep{DBLP:journals/corr/ChenPSA17}. It uses atrous convolution \citep{DBLP:journals/corr/YuK15,DBLP:journals/pami/ChenPKMY18}, which generates empty space between each element of the convolutional kernel. To extract both global and local information from a feature map, the ASPPHead adopts multiple atrous convolutions with various atrous rates in parallel. For the last feature map $\bC_5$, the first branch applies a series of global average pooling (GAP), convolutional unit block, and bilinear upsampling to restore the spatial size prior to the GAP. Each of the other four branches applies a convolutional unit block whose convolutional operation adopts an atrous rate $\{1, a, 2a, 3a\}$, where $a=96/s$. The results from the five branches are concatenated together with respect to channel dimension, and then a convolutional unit block $h$ is applied to fuse them. Denoting the outputs of convolutional unit blocks in parallel branches as $\{\bP_{\GAP}, \bP_1, \bP_a, \bP_{2a}, \bP_{3a}\}$, fusing them is represented as
\begin{align}
	\bZ = h([\UP_{H/s}(\bP_{\GAP}); \bP_1; \bP_a; \bP_{2a}; \bP_{3a}]).
\end{align}
Similarly, the fused feature $\bZ$ is then subjected to a $1\times1$ convolution and a $s\times$ bilinear upsampling to yield a predicted semantic mask $\hbY$, which has the size of $H \times W \times N_c$. In DeepLabV3+ \citep{DBLP:conf/eccv/ChenZPSA18}, a variant called SepASPPHead is developed using depthwise separable convolutions instead, while keeping the same decoder architecture. This single-stage feature fusion has also been used in other segmentation networks such as FCN \citep{DBLP:conf/cvpr/LongSD15} and U-Net \citep{DBLP:conf/miccai/RonnebergerFB15}, which progressively repeats fusion for two features with upsampling at each time.

\paragraph{Summary and Generalization} As reviewed above, modern decoders of segmentation networks perform multi- or single-stage feature fusion, which we collectively refer to as \emph{multi-level feature fusion}. Although each decoder has a distinct architecture, their feature fusions share a similar design pattern (\cref{fig:head}). Using single or multiple encoder features, certain operations are applied in parallel branches, and the convolutional unit block in the $i$th branch generates the $i$th feature map $\bP_i$ for $i \in \{1, \cdots, n_b\}$ for the number of branches $n_b$. Because the spatial size of each feature map $\bP_i$ differs, optional $r_i \times$ bilinear upsampling is needed to assure the same spatial size. For notational simplicity, $1 \times$ bilinear upsampling is defined as the identity operation. Because the set of encoder features for fusion includes a feature map that does not require bilinear upsampling, at least one branch exhibits the upsampling ratio $r_i=1$, whereas others use $r_i > 1$. The fused feature is now obtainable by concatenation with respect to channel dimension and a convolutional unit block $h$ as
\begin{align}
	\bZ = h([\UP_{r_1}(\bP_1); \cdots; \UP_{r_{n_b}}(\bP_{n_b})]).
\end{align}

\section{Scale Disequilibrium}
\label{sec:method}

\begin{figure*}[t]
	\begin{center}
		\centerline{\includegraphics[width=1.00\textwidth]{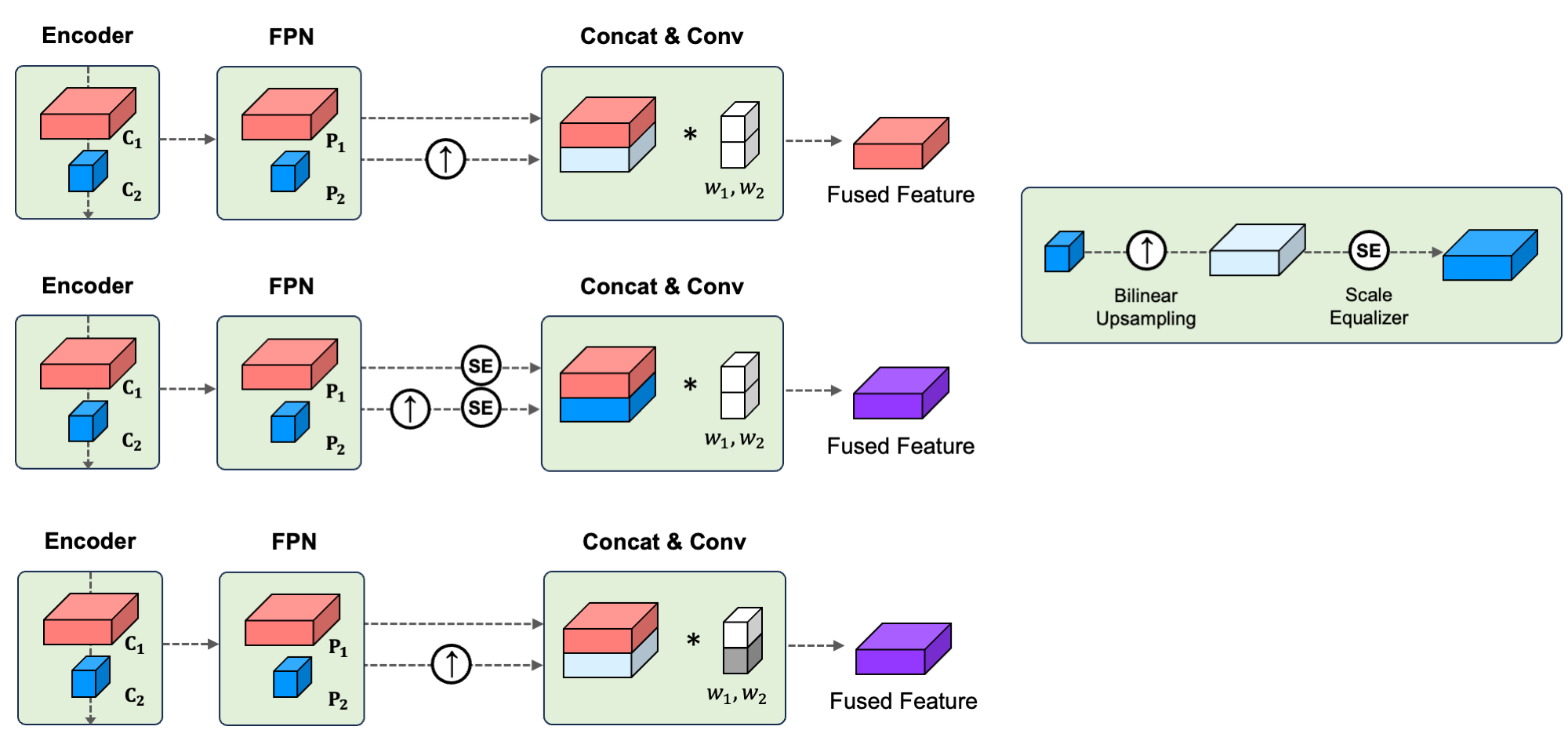}}
		\caption{
			Overview of the problem statement and the proposed solution. This illustration depicts a fusion by UPerHead for two features for simplicity, but nonetheless, the common fusion scheme uses four features. (Top) Existing multi-level feature fusion concatenates features after bilinear upsampling. The variances of concatenation subjects, represented as chroma in this figure, exhibit disequilibrium because bilinear upsampling decreases variance. In this fusion, $\bP_1$ dominates in the fused feature as a red color, which diminishes the contribution of $\bP_2$ and causes slower training on $w_2$. (Middle) Our proposed multi-level feature fusion with scale equalizers guarantees consistent variance across subjects of concatenation. In this scheme, a suitably fused feature as a purple color is produced with alive gradients with respect to both $w_1$ and $w_2$. (Bottom) Efficient implementation of our proposed method, where scale equalizers are replaced by applying auxiliary initialization for $w_1$ and $w_2$.
		}
		\label{fig:concept}
	\end{center}
	\vskip -0.2in
\end{figure*}

\subsection{Problem Statement}
\label{sec:problem}
As reviewed above, the decoder of the segmentation network includes a module to fuse features of varied sizes. Here, we claim that multi-level feature fusion requires explicit scale equalization because they exhibit different scales, which causes scale disequilibrium on gradients (\cref{fig:concept}).

To understand feature scale, this study uses feature variance. Other measures such as the norm depend on the size of the feature, whereas variance provides a suitably scaled result with respect to its size. Owing to the effectiveness of variance in understanding feature scales, it has been adopted in several pieces of literature \citep{DBLP:journals/jmlr/GlorotB10,DBLP:conf/iccv/HeZRS15,DBLP:conf/nips/KlambauerUMH17}. We also employ the mean of a feature to understand its representative value as occasion arises. Using variance, we describe the scale disequilibrium as follows.

\begin{proposition}
	\label{prop:diseq}
	Consider a multi-level feature fusion, where a concatenated feature $[x_1; x_2]$ is subjected to a linear layer with weight $[w_1, w_2]$ and bias $b$ to yield the fused feature $y=w_1 x_1 + w_2 x_2 + b$. When the two features $x_1$ and $x_2$ are on different scales, \ie, $\Var[x_1] \neq \Var[x_2]$, the gradients of the fused feature with respect to the corresponding weight exhibit scale disequilibrium, \ie, $\Var[\pdv{y}{w_1}] \neq \Var[\pdv{y}{w_2}]$.
\end{proposition}

The proof is straightforward because $\pdv{y}{w_i}=x_i$. From the chain rule, the gradient of a loss function $L$ with respect to weight $w_i$ is $\pdv{L}{w_i}=\sum_{y}\pdv{L}{y}\pdv{y}{w_i}$, and thus the gradient scale is affected by the scale of the corresponding feature $x_i$. The term linear layer indicates a fully connected layer or a convolutional layer.

For example, consider scale disequilibrium for concatenation subjects where $\Var[x_1] = 10 \Var[x_2]$. Then we obtain $\Var[\pdv{y}{w_1}] = 10 \Var[\pdv{y}{w_2}]$, and thus gradient descent on $w_2$ is on a ten times smaller scale than $w_1$, which slows down the training on $w_2$. However, gradient descent optimizers inherently assume scale equilibrium on gradients \citep{DBLP:journals/corr/abs-1212-5701}: For gradient descent $w_i \leftarrow w_i - \eta \pdv{L}{w_i}$, the weight initializer sets the same scale of initial weight $\Var[w_i]$ for weights within the same linear layer, and common gradient descent uses a single learning rate $\eta$ without scale discrimination, which leads to difficulty in capturing different gradient scales $\Var[\pdv{L}{w_i}]$. Note that existing literature \citep{DBLP:journals/jmlr/GlorotB10,DBLP:conf/iccv/HeZRS15,DBLP:conf/nips/KlambauerUMH17,DBLP:conf/uai/BachlechnerMMCM21} have discussed matching gradient scales across inter-layers to ensure stable gradient descent without poor training dynamics such as vanishing or exploding gradients. On top of inter-layer gradient scales, we claim to equalize intra-layer gradient scales. For the feature fusion scenario, matching the intra-layer gradient scales $\Var[\pdv{y}{w_1}] = \Var[\pdv{y}{w_2}]$ requires scale equalization for the subjects of concatenation: $\Var[x_1] = \Var[x_2]$. Achieving scale equilibrium eliminates the hidden factor that causes degradation in gradient descent optimization, which enhances the training of the segmentation network as well as the performance of semantic segmentation.

Furthermore, the gradient scale indicates the amount of contribution: A smaller scale on the gradient or feature indicates less contribution to the predicted mask. For example, when the last feature map that contains rich, high-level image information contributes little to the predicted mask, the quality of the segmentation result would degrade. To use the last feature map while supplementing its deficient information using multi-level features, it is desirable to ensure a balanced contribution from the multi-level features. Note that we are not saying that the amount of feature contributions should be precisely controlled to be optimal at initialization; rather, we would like to equalize feature contributions at the initial state and then let them change to be optimal during training. Our claim is that unwanted imbalances in gradient scales should be avoided at initialization. Indeed, the existing literature on matching inter-layer gradient scales have approached it this way and emphasized avoiding vanishing or exploding gradients at the initial state; thereafter, gradient scales are allowed to change during training. Considering this objective, our argument can be interpreted as establishing a valid initialization to achieve scale equilibrium on gradients with respect to intra-layer weights.

These arguments can be extended to match the mean of gradients, which requires the same mean for the subjects of concatenation: $\E[x_1] = \E[x_2]$. Based on this claim, we inspect the scale of concatenation subjects in the modern decoder of the segmentation network.

\paragraph{Batch Normalization Partially Equalizes Scale} Fortunately, the use of batch normalization results in a normalized feature\footnote{For batch normalization $\gamma \hat{x} + \beta$, the initial state where $\gamma=1$ and $\beta=0$ provides a normalized feature $\hat{x}$.} and thus concatenation of several features from the output of batch normalization is allowable. Moreover, batch normalization allows the use of convolution with arbitrary weights $\bW$ and ReLU operation without causing scale disequilibrium. This is because the output of a convolutional unit block with the pipeline [Conv--BatchNorm--ReLU] yields a fixed mean and variance without requiring any specific conditions on the weight $\bW$ and feature $\bx$:
\begin{align}
	\E[\ReLU(\BatchNorm(\bW\bx))]   & = \frac{1}{\sqrt{2 \pi}}, \\
	\Var[\ReLU(\BatchNorm(\bW\bx))] & = \frac{\pi - 1}{2 \pi}.
\end{align}
This property also implies that any architecture can be freely chosen before the input of the convolutional unit block. Furthermore, batch normalization guarantees a consistent mean and variance for each channel \citep{DBLP:conf/icml/IoffeS15}. This channel-wise normalization is preferable because the output features from multiple branches are concatenated with respect to channel dimension. These characteristics of batch normalization explain why it is still preferred for the decoder of segmentation networks, despite the existence of several alternatives, such as layer normalization, which does not perform channel-wise normalization \citep{DBLP:journals/corr/BaKH16}. In summary, batch normalization provides a feature with a consistent scale, which allows the concatenation of several features from convolutional unit blocks.

\paragraph{Bilinear Upsampling Breaks Scale Equilibrium} However, even with batch normalization, feature scales exhibit disequilibrium when subsequently using bilinear upsampling. Consider a multi-level feature fusion for $\{\bP_1, \bP_2\}$, where each feature is an output of a convolutional unit block, and the latter $\bP_2$ needs $r \times$ bilinear upsampling with $r>1$ to become the same spatial size as $\bP_1$. Fusing them requires computing
\begin{align}
	\bZ^{\fuse} = \bW^{\fuse}[\bP_1; \UP_{r}(\bP_2)],
\end{align}
which is an intermediate result after convolutional layer of fusion with $\bW^{\fuse}$. Although convolutional unit blocks on parallel branches assure consistent scales for $\{\bP_1, \bP_2\}$, scales of concatenation subjects $\{\bP_1, \UP_{r}(\bP_2)\}$ are not guaranteed to be equal. Indeed, we claim that scale disequilibrium occurs during this feature fusion due to bilinear upsampling. Specifically, we prove that bilinear upsampling decreases feature variance:
\begin{theorem}
	\label{thm:bilinear}
	Bilinear upsampling decreases feature variance, \ie, $\Var[\UP_{r}(\bX)] < \Var[\bX]$ for upsampling ratio $r>1$ and feature $\bX$ that is not a constant feature.
\end{theorem}
The constant feature here indicates a vector with the same constant elements. Note that bilinear upsampling conserves feature mean---but not feature variance. Furthermore, variance provides a suitably scaled result with respect to its size, which ensures that the decreased variance is not caused by the increased size due to upsampling. The decreased variance is caused by the linear interpolation function used in bilinear upsampling, which does not conserve the second moment that is included in the variance. See the Appendix for a detailed proof and further discussion.

In \cref{sec:back}, we reviewed multi-level feature fusion in modern decoders and found that, as a general rule, at least one branch uses the upsampling ratio $r_i=1$, whereas others show $r_i > 1$. Therefore, \cref{thm:bilinear} indicates that modern decoders with multi-level feature fusion exhibit scale disequilibrium. The fatal problem is that the last feature map always requires bilinear upsampling, which reduces its feature and gradient scales, obstructing the use of its rich information on an image. This problem arises even when using batch normalization: Because bilinear upsampling is applied after each convolutional unit block, the equalized scales subsequently change.

\begin{figure}[t]
	\begin{center}
		\centerline{\includegraphics[width=1.00\columnwidth]{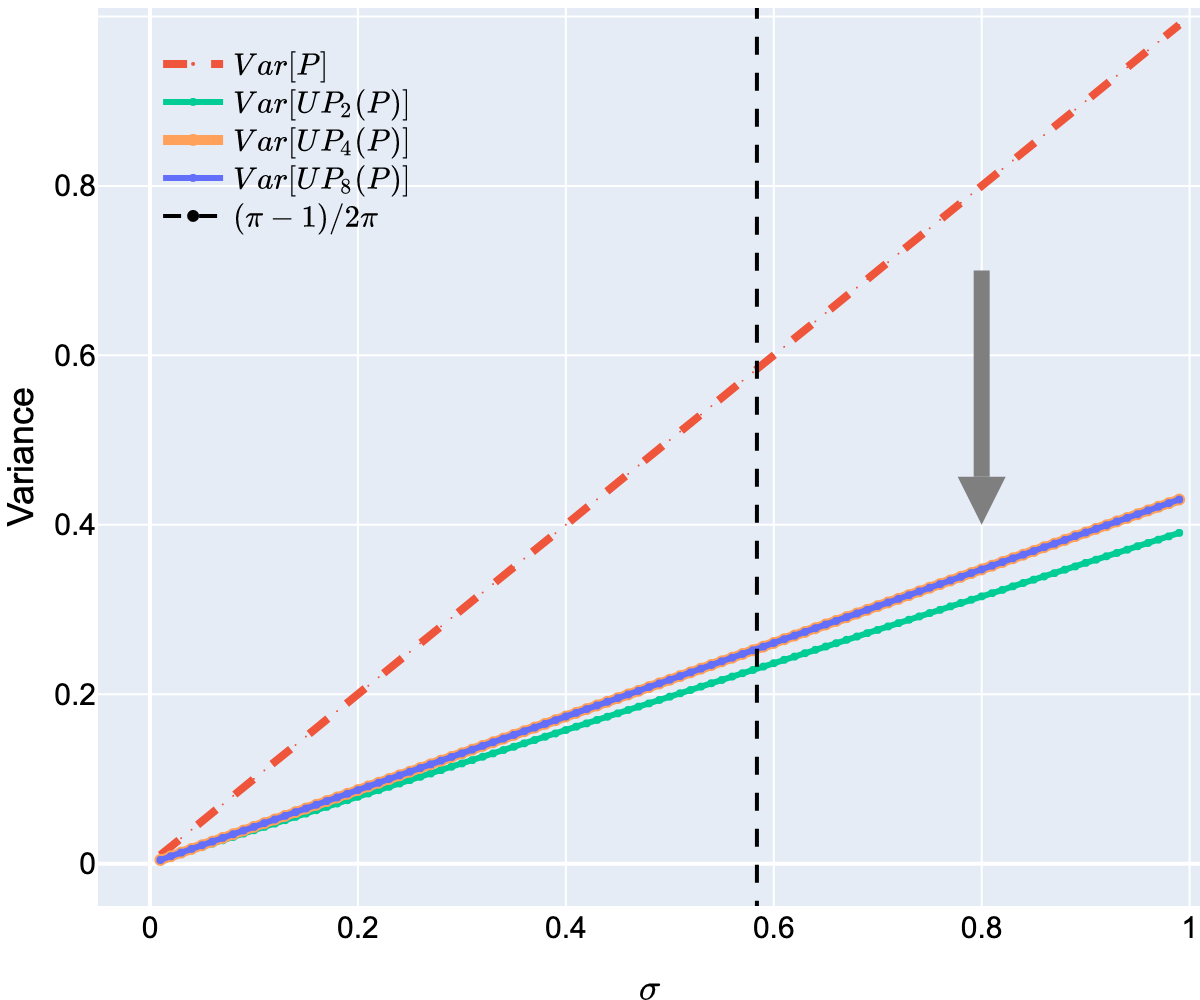}}
		\caption{
			Empirical observation on decreased variance after bilinear upsampling. The black dotted line $(\pi-1)/2\pi$ corresponds to the case when the output of a convolutional unit block is subjected to bilinear upsampling.
		}
		\label{fig:emp}
	\end{center}
	\vskip -0.2in
\end{figure}

\paragraph{Empirical Observation} Now, we empirically demonstrate decreased variance after bilinear upsampling. Considering a practical feature fusion scenario, we generated artificial random normal data $\bP$ sampled from $\N(1/\sqrt{2 \pi}, \sigma^2)$, which corresponds to a feature after a convolutional unit block but before bilinear upsampling. The feature $\bP$ is set to have width 128, height 128, number of channels 256, and mini-batch size 16. Then we applied $r \times$ bilinear upsampling to $\bP$ with $r \in \{2, 4, 8\}$ and measured the variance of each outcome. \figref{fig:emp} summarizes the results across various choices of $\sigma \in (0, 1)$. We observed that bilinear upsampling decreased feature variance in all simulations.

\subsection{Proposed Solution: Scale Equalizer}
Our claim is that we should modify the existing feature fusion method to achieve scale equilibrium for concatenation subjects---the output of each branch that ends with bilinear upsampling. This objective may be accomplished in several ways. The naive approach is to place batch normalization after bilinear upsampling, changing the pipeline from [Conv--BatchNorm--ReLU--UP] to [Conv--ReLU--UP--BatchNorm]. This pipeline yields a normalized feature with a consistent scale but requires extra computational cost. Because batch normalization computes the mean and standard deviation (std) of the current incoming feature map across the mini-batch, its computational complexity increases with the larger size of the feature \citep{DBLP:conf/cvpr/HuangYLD18}. The computational complexity of the backward operation further increases with the larger size of the feature map because the derivative for batch normalization is much more complicated \citep{DBLP:conf/cvpr/Yao0ZHL21}. Consequently, applying batch normalization to an upsampled feature causes a significantly more expensive computation compared with that of a non-upsampled feature. Considering this problem, we alternatively explore a computationally efficient solution to acquire scale equilibrium.

Here, we propose \emph{scale equalizer}, a module to be injected after bilinear upsampling but before concatenation. To achieve scale equilibrium at minimal cost, we design the scale equalizer as simple as possible. Specifically, our proposed scale equalizer normalizes target feature $\bx$ using global mean $\mu$ and global std $\sigma$ as $\ScaleEqualizer(\bx) \coloneqq (\bx-\mu)/\sigma$. The global mean and std are scalars computed from the target feature $\bx$ across the training dataset, which can be performed before training. Once the global mean and std are obtained, they can be set as fixed constants during training, which simplifies forward and backward operations for the scale equalizer. By contrast, mean and std are not constants for common normalization operations such as batch normalization or layer normalization because they use a mean and std of a current incoming feature. Thus, compared with existing normalization operations, the proposed scale equalizer can be implemented with little extra cost.

\paragraph{Scale Equalizers Equalize Scales}
Now consider multi-level feature fusion with scale equalizers, where the scale equalizer is applied after bilinear upsampling of each branch but before concatenation. The concatenation subject $\ScaleEqualizer_i(\UP_{r_i}(\bP_i))$ exhibits zero mean and unit variance, which assures scale equilibrium. Because the scale equalizer uses empirically measured values of the global mean and std, the scale equilibrium does not require architectural restrictions or specific conditions on weight. In other words, scale equilibrium is always guaranteed for any dataset and any architecture of segmentation network.

\paragraph{Efficient Implementation via Initialization}
In fact, matching intra-layer gradient scales via scale equalizers can be interpreted as establishing a valid initialization. For multi-level feature fusion $y=\sum_i w_i x_i + b$, after replacing $x_i$ with $\ScaleEqualizer_i(x_i) = (x-\mu_i)/\sigma_i$, we obtain
\begin{align}
	y & = \sum_i{\left(\frac{w_i}{\sigma_i}\right)x_i} + \left(b-\sum_i{\frac{w_i \mu_i}{\sigma_i}}\right).
\end{align}
Thus, injecting scale equalizers is equivalent to adopting an auxiliary initializer with $w^{\prime}_i = w_i / \sigma_i$ and $b^{\prime} = b-\sum_i{w_i \mu_i / \sigma_i}$. This auxiliary initializer means calibrating the weights and bias in the linear layer of fusion in advance using expected feature scales. Furthermore, because batch normalization follows subsequently (\cref{sec:back}), the latter for bias correction is actually not needed, whereas the former for weight calibration is still needed to control the scales of concatenation subjects. For UPerHead as a concrete example, weights in the convolutional layer of fusion are partitioned into four groups with respect to channel dimension, and the weights in each group $w_i$ are re-scaled via the global std $\sigma_i$. In summary, after primary initialization of the decoder, we compute the global mean and std for each target feature, apply the auxiliary initializer to weights, and then proceed with the main training (\cref{alg:aux}). This implementation requires no additional computational cost during main training, which enables us to achieve scale equilibrium for free.

\begin{algorithm}[tb]
	\caption{Efficient Implementation via Initialization}
	\label{alg:aux}
	\begin{algorithmic}
		\STATE {\bfseries Input:} set of training images $S$, encoder $e$, decoder $d$
		\STATE Using pretrained weights $\Theta$, initialize encoder $e$.
		\STATE Using preferred initializers, initialize decoder $d$ into weight $\Omega$, including $\{w_i\}$.
		\STATE Set $m_{1,i} = m_{2,i} = 0$ for $i \in \{1, \cdots, n_b\}$.
		\FOR{$\bI \in S$}
		\STATE $\{\bC_i\} = e_{\Theta}(\bI)$
		\FOR{$i=1$ {\bfseries to} $n_b$}
		\STATE $m_{1,i} = m_{1,i} + \E[\bC_i]$
		\STATE $m_{2,i} = m_{2,i} + \E[\bC_i^2]$
		\ENDFOR
		\ENDFOR
		\FOR{$i=1$ {\bfseries to} $n_b$}
		\STATE Obtain the global mean $\mu_i = m_{1,i}/|S|$.
		\STATE Obtain the global std $\sigma_i = \sqrt{m_{2,i}/|S|-\mu_i^2}$.
		\STATE Update $w_i$ in $\Omega$ using auxiliary initializer $w_i^{\prime} = w_i / \sigma_i$.
		\ENDFOR
		\STATE Using the updated decoder weight $\Omega^{\prime}$, run the main training for encoder $e_{\Theta}$ and decoder $d_{\Omega^{\prime}}$.
	\end{algorithmic}
\end{algorithm}

\section{Experiments}
\label{sec:exp}

\begin{table*}[t]
	\caption{Summarization of mIoU (\%) from semantic segmentation experiments with multi-stage feature fusion using UPerHead. ``Scale EQ'' indicates the scale equalizers, and ``Diff'' indicates the mIoU difference after injecting the scale equalizers.}
	\label{tab:multi}
	\begin{center}
		\begin{small}
			\begin{sc}
				\resizebox{1.0\textwidth}{!}{
					\begin{tabular}{l|ccc|ccc}
						\toprule
						Dataset                                         & \multicolumn{3}{c|}{ADE20K} & \multicolumn{3}{c}{PASCAL VOC 2012 AUG}                                                \\
						\midrule
						Encoder                                         & w/o Scale EQ                & w/ Scale EQ                             & Diff   & w/o Scale EQ & w/ Scale EQ & Diff   \\
						\midrule
						Swin-T  \citep{DBLP:conf/iccv/LiuL00W0LG21}     & 43.384                      & 43.576                                  & +0.192 & 78.750       & 78.996      & +0.246 \\
						Swin-S                                          & 47.298                      & 47.486                                  & +0.188 & 81.940       & 82.138      & +0.198 \\
						Swin-B                                          & 47.490                      & 47.648                                  & +0.158 & 82.200       & 82.378      & +0.178 \\
						Twins-SVT-S \citep{DBLP:conf/nips/ChuTWZRWXS21} & 44.914                      & 45.018                                  & +0.104 & 80.448       & 80.732      & +0.284 \\
						Twins-SVT-B                                     & 47.224                      & 47.500                                  & +0.276 & 82.048       & 82.524      & +0.476 \\
						Twins-SVT-L                                     & 48.648                      & 48.894                                  & +0.246 & 82.168       & 82.404      & +0.236 \\
						ConvNeXt-T \citep{DBLP:conf/cvpr/0003MWFDX22}   & 45.024                      & 45.300                                  & +0.276 & 80.668       & 80.932      & +0.264 \\
						ConvNeXt-S                                      & 47.736                      & 47.866                                  & +0.130 & 82.472       & 82.650      & +0.178 \\
						ConvNeXt-B                                      & 48.376                      & 48.684                                  & +0.308 & 82.934       & 83.038      & +0.104 \\
						\bottomrule
					\end{tabular}
				}
			\end{sc}
		\end{small}
	\end{center}
	\vskip -0.1in
\end{table*}

\begin{table*}[t]
	\caption{Summarization of mIoU (\%) from semantic segmentation experiments with single-stage feature fusion using various heads.}
	\label{tab:single}
	\begin{center}
		\begin{small}
			\begin{sc}
				\resizebox{1.0\textwidth}{!}{
					\begin{tabular}{l|ccc|ccc}
						\toprule
						Dataset                                       & \multicolumn{3}{c|}{Cityscapes} & \multicolumn{3}{c}{ADE20K}                                                \\
						\midrule
						Decoder                                       & w/o Scale EQ                    & w/ Scale EQ                & Diff   & w/o Scale EQ & w/ Scale EQ & Diff   \\
						\midrule
						FCNHead     \citep{DBLP:conf/cvpr/LongSD15}   & 76.578                          & 76.972                     & +0.394 & 39.780       & 39.958      & +0.178 \\
						PSPHead \citep{DBLP:conf/cvpr/ZhaoSQWJ17}     & 79.394                          & 79.858                     & +0.464 & 43.970       & 44.228      & +0.258 \\
						ASPPHead \citep{DBLP:journals/corr/ChenPSA17} & 79.312                          & 79.720                     & +0.408 & 44.854       & 45.004      & +0.150 \\
						SepASPPHead \citep{DBLP:conf/eccv/ChenZPSA18} & 80.448                          & 80.592                     & +0.144 & 45.144       & 45.486      & +0.342 \\
						\bottomrule
					\end{tabular}
				}
			\end{sc}
		\end{small}
	\end{center}
	\vskip -0.1in
\end{table*}

\subsection{Multi-Stage Feature Fusion}
\label{sec:expmulti}
\paragraph{Objective} So far, we have discussed the need for scale equalizers for multi-level feature fusion. The objective here is to compare the segmentation performance before and after injecting scale equalizers into multi-stage feature fusion. We considered extensive setups, such as the choice of backbone and target dataset. For the backbone network, we employed recent vision transformers that achieved state-of-the-art performance. Nine backbones of Swin-\{T, S, B\} \citep{DBLP:conf/iccv/LiuL00W0LG21}, Twins-SVT-\{S, B, L\} \citep{DBLP:conf/nips/ChuTWZRWXS21}, and ConvNeXt-\{T, S, B\} \citep{DBLP:conf/cvpr/0003MWFDX22} pretrained on ImageNet-1K \citep{DBLP:conf/cvpr/DengDSLL009} were examined, where T, S, B, and L stand for tiny, small, base, and large models, respectively. These encoders require bilinear upsampling for multi-stage feature fusion. Targeting multi-stage feature fusion, we employed UPerHead \citep{DBLP:conf/eccv/XiaoLZJS18}. Two datasets were examined, including the ADE20K \citep{DBLP:journals/ijcv/ZhouZPXFBT19} and PASCAL VOC 2012 \citep{DBLP:journals/ijcv/EveringhamEGWWZ15}.

\paragraph{Hyperparameters} To follow common practice for semantic segmentation, training recipes from \texttt{MMSegmentation} \citep{mmseg2020} were employed. For training with Swin and Twins encoders, AdamW optimizer \citep{DBLP:conf/iclr/LoshchilovH19} with weight decay $10^{-2}$, betas $\beta_1=0.9, \beta_2=0.999$, and learning rate $6 \times 10^{-5}$ with polynomial decay of the 160K scheduler after linear warmup were used. For training with ConvNeXt encoders, AdamW optimizer with weight decay $5 \times 10^{-2}$, betas $\beta_1=0.9, \beta_2=0.999$, learning rate $10^{-4}$ with polynomial decay of the 160K scheduler after linear warmup, and mixed precision training \citep{DBLP:conf/iclr/MicikeviciusNAD18} were used. The training was conducted on a 4$\times$ GPU machine, and SyncBN \citep{DBLP:conf/cvpr/0005DSZWTA18} was used for distributed training. We measured the mean intersection over union (mIoU) and reported the average of five runs with different random seeds.

\paragraph{Datasets} The ADE20K dataset contains scene-centric images along with the corresponding segmentation labels. A crop size of $512 \times 512$ pixels was used, which was obtained after applying mean-std normalization and a random resize operation using a size of $2048 \times 512$ pixels with a ratio range of 0.5 to 2.0. Furthermore, a random flipping with a probability of 0.5 and the photometric distortions were applied. The objective was to classify each pixel into one of the 150 categories and train the segmentation network using the pixel-wise cross-entropy loss. The same goes for the PASCAL VOC 2012 dataset with 21 categories, and we followed the augmented PASCAL VOC 2012 dataset.

\paragraph{Results} We observed that injecting scale equalizers into multi-stage feature fusion improved the mIoU index (\tabref{tab:multi}). The mIoU increases of about +0.1 to +0.3 were consistently observed across all setups of nine backbones and two datasets.

\subsection{Single-Stage Feature Fusion}

\paragraph{Objective} Now we examine single-stage feature fusion. The target encoder was ResNet-101 \citep{DBLP:conf/cvpr/HeZRS16}, which was modified to exhibit output stride $s=8$ and was pretrained on ImageNet-1K. Various decoders were examined, including FCNHead \citep{DBLP:conf/cvpr/LongSD15}, PSPHead \citep{DBLP:conf/cvpr/ZhaoSQWJ17}, ASPPHead \citep{DBLP:journals/corr/ChenPSA17}, and SepASPPHead \citep{DBLP:conf/eccv/ChenZPSA18}. The target datasets were the Cityscapes \citep{DBLP:conf/cvpr/CordtsORREBFRS16} and ADE20K datasets.

\paragraph{Hyperparameters} Similar to \cref{sec:expmulti}, training recipes from \texttt{MMSegmentation} were employed. For training on the Cityscapes dataset, stochastic gradient descent with momentum 0.9, weight decay $5 \times 10^{-4}$, and learning rate $10^{-2}$ with polynomial decay of the 80K scheduler were used. The same goes for training on the ADE20K dataset while using the 160K scheduler instead. The training was conducted on a 4$\times$ GPU machine, and SyncBN was used for distributed training. We measured the mIoU and reported the average of five runs with different random seeds.

\paragraph{Datasets} The Cityscapes dataset contains images of urban street scenes along with the corresponding segmentation labels. A crop size of $1024 \times 512$ pixels was used, which was obtained after applying mean-std normalization and a random resize operation using a size of $2048 \times 1024$ pixels with a ratio range of 0.5 to 2.0. Furthermore, a random flipping with a probability of 0.5 and the photometric distortions were applied. The objective was to classify each pixel into one of the 19 categories and train the segmentation network using the pixel-wise cross-entropy loss. Experiments on the ADE20K followed the description in \cref{sec:expmulti}.

\paragraph{Results} Similarly, injecting scale equalizers consistently improved the mIoU index across all setups of four decoder heads and two datasets (\tabref{tab:single}), which verifies the effectiveness of scale equalizers for any choice of architecture. See the Appendix for more experimental results.

\section{Conclusion}
\label{sec:con}
This study discussed the scale disequilibrium in multi-level feature fusion. First, we reviewed the mechanisms of existing segmentation networks, which perform multi- or single-stage feature fusion. We demonstrated that the existing multi-level feature fusion exhibits scale disequilibrium due to bilinear upsampling, which causes degraded gradient descent optimization. To address this problem, injecting scale equalizers is proposed to guarantee scale equilibrium for multi-level feature fusion. Experiments showed that the use of scale equalizers consistently increased the mIoU index by about +0.1 to +0.4 across numerous datasets and architectures. We hope that our proposed problem and solution will facilitate the research community in developing an improved multi-level feature fusion and segmentation network.

\bibliography{example_paper}
\bibliographystyle{icml2024}


\end{document}